\documentclass{article}

\PassOptionsToPackage{numbers, compress}{natbib}
\usepackage[preprint]{neurips_2021}
\usepackage{dsfont}

\usepackage[utf8]{inputenc} 
\usepackage[T1]{fontenc}    
\usepackage{hyperref}       
\usepackage{url}            
\usepackage{booktabs}       
\usepackage{amsfonts}       
\usepackage{nicefrac}       
\usepackage{microtype}      
\usepackage{xcolor}         
\usepackage{amsmath}
\usepackage{natbib}
\usepackage{adjustbox}
\usepackage{graphicx}
\usepackage{caption}
\usepackage{subcaption}

\newcommand{\tb}[1]{\textbf{#1}}

\title{Simplicial Attention  Neural  Networks}

\author{%
  Lorenzo Giusti\thanks{Equal contribution} \\
  DIAG Department \\
  Sapienza University or Rome\\
  Via Ariosto 25, 00185, Rome, Italy \\
  \texttt{lorenzo.giusti@uniroma1.it} \\
  \And
  Claudio Battiloro$^{*}$ \\
  DIET Department \\
  Sapienza University or Rome\\
  Via Eudossiana 18, 00184, Rome, Italy \\
  \texttt{claudio.battiloro@uniroma1.it} \\
  \And
  Paolo Di Lorenzo \\
  DIET Department \\
  Sapienza University or Rome\\
  Via Eudossiana 18, 00184, Rome, Italy \\
  \texttt{paolo.dilorenzo@uniroma1.it} \\
  \And
  Stefania Sardellitti \\
  DIET Department \\
  Sapienza University or Rome\\
  Via Eudossiana 18, 00184, Rome, Italy \\
  \texttt{stefania.sardellitti@uniroma1.it} \\
  \And
  Sergio Barbarossa \\
  DIET Department \\
  Sapienza University or Rome\\
  Via Eudossiana 18, 00184, Rome, Italy \\
  \texttt{sergio.barbarossa@uniroma1.it}
}

\begin{document}

\maketitle

\vspace{-.3cm}
\begin{abstract}
The aim of this work is to introduce Simplicial Attention Neural Networks (SANs), i.e., novel neural architectures that operate on data defined on simplicial complexes leveraging masked self-attentional layers.
Hinging on formal arguments from topological signal processing, we introduce a proper self-attention mechanism able to process data components at different layers (e.g., nodes, edges, triangles, and so on), while learning how to weight both upper and lower neighborhoods of the given topological domain in a  task-oriented fashion.  
The proposed SANs generalize most of the current architectures available for processing data defined on simplicial complexes. The proposed approach compares favorably with other methods when applied to different (inductive and transductive) tasks such as trajectory prediction and missing data imputations in citation complexes.
\end{abstract}

\section{Introduction}
In the last few years, the fast and wide development of deep learning techniques dramatically improved the state-of-the-art in many learning tasks. From Feed-Forward \cite{goodfellow2016deep} to Transformer \cite{vaswani2017attention}, passing for Convolutional \cite{lecun2015deep} and Recurrent \cite{rumelhart1985learning} Neural Networks, more and more sophisticated architectures promoted huge advances from a both theoretical and practical perspective. Nowadays, data defined on irregular domains (e.g, graphs) are pervasive, with applications ranging from social networks, recommender systems, cyber-security, sensor networks, and natural language processing. Since their introduction \cite{scarselli2008graph, gori2005new}, Graph Neural Networks (GNNs) have shown remarkable results in learning tasks with data defined over a graph domain, where the flexibility of neural networks is coupled with prior knowledge about data relationships, expressed in terms of a graph topology. The literature on GNNs is huge and various approaches have been explored, mainly grouped in spectral \cite{Bruna19, kipf2016semi} and non-spectral methods \cite{hamilton2017inductive, DuvenaudMABHAA15, AtwoodT16}. Generally speaking, the idea is learning from data defined over graphs by computing a principled representation of node features via local aggregation with the information gathered from neighbors, defined by the underlying graph topology. Exploiting this simple, but powerful, idea, outstanding performance has been reached in many tasks such as node or graph classification  \cite{kipf2016semi, velivckovic2017graph, hamilton2017inductive} and link prediction \cite{zhang2018link}, just to name a few. \\
At the same time, a significant performance boost to deep learning techniques has been given by the introduction of attention mechanisms. They have been firstly introduced to handle sequence-based tasks \cite{bahdanau2014neural, gehring2016convolutional}, allowing for variable sized inputs, to focus on the most relevant parts of them. Then, pioneering works have generalized attention mechanisms to data defined over graphs \cite{velivckovic2017graph, yun2019graph}.  Nowadays, attention-based models (including Transformers) have a huge range of shades and applications spanning from learning sentence representations \cite{Lin19} to machine translation \cite{bahdanau2014neural}, from machine reading  \cite{Cheng20} to multi-label image classification \cite{Chen19}, both on euclidean and graph domains, and they achieve state-of-the-art results in many of these tasks. \\
However, despite their overwhelming popularity, graph-based representations can only take into account pairwise interactions. As a result, graphs are not always able to capture all the information present in complex interconnected systems, where the interactions cannot be reduced to simple pairwise relationships. In biological networks, for example, multi-way links among complex substances, such as genes, proteins, or metabolites, cannot be evoked using simply pairwise relationships \cite{lambiotte2019networks}. 
Very recent works on topological signal processing \cite{barbarossa2020topological,schaub2021signal} demonstrated the benefit of learning from data defined on simplicial complexes, which are specific examples of hyper-graphs with a rich algebraic description, and can easily encode multi-way relationships hidden in the data. Consequently, there was a natural interest in the development of (deep) neural networks architectures able to handle data defined on simplicial complexes, as illustrated in the sequel.\\
\textbf{Related Works.} Simplicial deep learning is a very recent and challenging research area. In \cite{ebli2020simplicial}, authors introduced a simplicial neural network architecture that generalizes graph convolutional networks (GCNs) exploiting higher-order Laplacians, without however considering separate processing for the lower and upper neighbourhoods of a simplicial complex. In \cite{bodnar2021weisfeiler}, message passing neural networks (MPNNs) \cite{gilmer2017neural} are adapted to simplicial complexes, with the aggregation and updating functions able to process the data exploiting lower and upper neighbourhoods. The architecture in \cite{bodnar2021weisfeiler} can be seen also as a generalization of the architectures in \cite{bunch2020simplicial, roddenberry2021principled} with a specific aggregation function given by simplicial filters \cite{yang2021finite}. In \cite{roddenberry2019hodgenet}, recurrent MPNNs architectures are considered for flow interpolation and graph classification tasks.
The work in \cite{yang2021simplicial} explicitly enables multihop processing based on upper and lower neighbourhoods, offering also spectral interpretability through the definition of simplicial filters and simplicial fourier transform \cite{yang2021finite}. None of these works considered attention mechanism in simplicial neural networks. 
Then, at the same time of our work, the paper in \cite{anonymous2022SAT} independently proposed an attention mechanism for simplicial neural networks. However, as we will clarify in the sequel, our approach differs from the one in \cite{anonymous2022SAT} in some key aspects. \\
\textbf{Contribution.} The aim of this paper is to introduce simplicial attention neural networks, i.e., a novel topological neural architecture that exploits self-attention mechanisms to process data defined on simplicial complexes. The architecture processes simplicial data via \textit{convolutional filtering operations} that take into account both the lower and upper neighbourhoods defined by the underlying topology. Moreover, we introduce \textit{a proper technique to process the harmonic component of the data} based on the design of a sparse projection operator. We illustrate how the proposed architecture generalizes almost all of the current architectures available for the processing of data defined on simplicial complexes. Although the work in \cite{anonymous2022SAT} proposes an ideally similar attention mechanism, crucial differences exist with our work. First, the work in \cite{anonymous2022SAT} proposes a MPNN architecture, while we consider a convolutional architecture with theoretical interpretation given by topological signal processing arguments. Second, our architecture allows for a tailored processing of the harmonic  component of the data, which cannot be derived from the architecture in \cite{anonymous2022SAT}. \textcolor{black}{Last, the method in \cite{anonymous2022SAT} sets up a single attention function to learn the attention coefficients over both lower and upper neighborhoods (cf. (2) and (3) in \cite{anonymous2022SAT}), whereas we develop two different attention mechanisms that work independently over the two domains.} 
To assess the effectiveness of our SANs with the other available architectures, we consider two challenging tasks: trajectory prediction on ocean drifters tracks data \cite{bodnar2021weisfeiler} (inductive learning) and missing data imputation in citation complexes \cite{ebli2020simplicial,yang2021simplicial} (transductive learning), fairly enhancing the boost given by self-attention and harmonic projection. We show that the proposed architecture performs favorably with respect to the current state-of-the-art. 




\vspace{-.3cm}
\section{Background on Topological Deep Learning}
\vspace{-.3cm}

In this section, we review some concepts from topological signal processing and deep learning that will be useful to introduce the proposed SAN architecture.\smallskip\\ 
\noindent \textbf{Simplicial complex and signals:} Given a finite set of vertices $\mathcal{V}$, a $k$-simplex $\mathcal{H}_{k}$ is a subset of $\mathcal{V}$ with cardinality $k+1$. A face of $\mathcal{H}_{k}$ is a subset with cardinality $k$ and thus a k-simplex has $k+1$ faces. A coface of $\mathcal{H}_{k}$ is a $(k + 1)$-simplex
that includes $\mathcal{H}_{k}$ \cite{barbarossa2020topological, lim2020hodge}. If two simplices share a common face, then they are lower neighbours; if they share a common coface, they are upper neighbours \cite{yang2021finite}.  A simplicial complex $\mathcal{X}_{k}$ of order $K$, is a collection of $k$-simplices $\mathcal{H}_{k}$, $k = 0, \ldots, K$ such that, for any $\mathcal{H}_{k} \in \mathcal{X}_{k}$, $\mathcal{H}_{k-1} \in \mathcal{X}_{k}$ if $\mathcal{H}_{k-1} \subset \mathcal{H}_{k}$ (inclusivity property). We denote the  set of k-simplex in $\mathcal{X}_{k}$ as  ${\cal D}_{k} := \{\mathcal{H}_{k}: \mathcal{H}_{k} \in \mathcal{X}_{k}\} $, with $|{\cal D}_{k}| = N_k$ and, obviously, ${\cal D}_{k} \subset {\cal X}_{k}$. We are interested in processing signals defined over a simplicial complex. A $k$-simplicial signal is defined as a mapping from the set of all $k$-simplices contained in the complex to real numbers:
\begin{equation}
\mathbf{x}_{k}: {\cal D}_{k} \rightarrow \mathbb{R}, \,\,\quad k=0, 1, \ldots K.
\end{equation}
The order of the signal is one less the cardinality of the elements of ${\cal D}_{k}$. In most of the cases the focus is on complex $\mathcal{X}_{2}$ of order up to two, thus a set of vertices $\mathcal{V}$ with $|\mathcal{V}| = V$, a set of edges $\mathcal{E}$ with $|\mathcal{E}|=E$ and a set of triangles $\mathcal{T}$ with $|\mathcal{T}| = T$ are considered, resulting in ${\cal D}_{0}={\cal V}$ (simplices of order 0), ${\cal D}_{1}={\cal E}$ (simplices of order 1) and ${\cal D}_{2}={\cal T}$ (simplices of order 2).


\textbf{Algebraic representations:} The structure of a simplicial complex ${\cal X}_{k}$  is fully described by the set of its incidence matrices $\mathbf{B}_{k}$, $k=1, \ldots, K$, given a reference orientation. The entries of the incidence matrix $\mathbf{B}_{k}$ establish which $k$-simplices are incident to which $(k-1)$-simplices.  Denoting the fact that two simplex have the same orientation with ${H}_{k-1,i} \sim {H}_{k,j}$ and viceversa with ${H}_{k-1,i} \not\sim {H}_{k,j},$ the entries of $\mathbf{B}_{k}$ are defined as follows:
  \begin{equation} \label{inc_coeff}
  \big[\mathbf{B}_{k} \big]_{i,j}=\left\{\begin{array}{rll}
  0, & \text{if} \; \mathcal{H}_{k-1,i} \not\subset \mathcal{H}_{k,j} \\
  1,& \text{if} \; \mathcal{H}_{k-1,i} \subset \mathcal{H}_{k,j} \;  \text{and} \; \mathcal{H}_{k-1,i} \sim \mathcal{H}_{k,j}\\
  -1,& \text{if} \; \mathcal{H}_{k-1,i} \subset \mathcal{H}_{k,j} \;  \text{and} \; \mathcal{H}_{k-1,i} \not\sim \mathcal{H}_{k,j}\\
  \end{array}\right. .
  \end{equation}
As an example, considering a simplicial complex $\mathcal{X}_{2}$ of order two, we have two incidence matrices $\mathbf{B}_{1} \in \mathbb{R}^{V \times E}$ and  $\mathbf{B}_{2} \in \mathbb{R}^{E \times T}$. From the incidence information, we can build the high order combinatorial Laplacian matrices \cite{goldberg2002combinatorial}, of order $k=0, \ldots, K$, as follows:
\begin{align}
&\mathbf{L}_{0}=\mathbf{B}_{1}\mathbf{B}_{1}^T,\label{Laplacian0}\\
&\mathbf{L}_{k}=\underbrace{\mathbf{B}_k^{T}\mathbf{B}_{k}}_{\mathbf{L}_k^{(d)}}+\underbrace{\mathbf{B}_{k+1}\mathbf{B}_{k+1}^T}_{\mathbf{L}_k^{(u)}}, \, k=1, \ldots, K-1, \label{Laplaciank}\\
&\mathbf{L}_{K}=\mathbf{B}_{K}^T\mathbf{B}_{K}.\label{LaplacianK}
\end{align}
All Laplacian matrices of intermediate order, i.e. $k=1, \ldots, K-1$, contain two terms: The first term $\mathbf{L}^{(d)}_k$, also known as  lower Laplacian, encodes the lower adjacency of $k$-order simplices; the second term $\mathbf{L}_k^{(u)}$, also known as upper Laplacian, encodes the upper adjacency of $k$-order simplices. Thus, for example, two edges are lower adjacent if they share a common vertex, whereas they are upper adjacent if they are faces of a common triangle. Note that the vertices of a graph can only be upper adjacent, if they are incident to the same edge. This is why the graph Laplacian $\mathbf{L}^{(0)}$ contains only one term, and it corresponds to the usual graph Laplacian.

\textbf{Hodge decomposition:}
High order Laplacians admit a Hodge decomposition \cite{lim2020hodge},
leading to three orthogonal subspaces. In particular, the $k$-simplicial signal space can be decomposed as:
\begin{equation} \label{hodge_spaces}
\mathbb{R}^{N_{k}} = \textbf{im}\big(\mathbf{B}_{k}^T\big) \bigoplus \textbf{im}\big(\mathbf{B}_{k+1}\big) \bigoplus \textbf{ker}\big(\mathbf{L}_{k}\big),
\end{equation}
where $\bigoplus$ is the direct sum of vector spaces and \textbf{ker}(·) and \textbf{im}(·) are the kernel and image spaces of a matrix, respectively. Thus, any signal $\mathbf{x}_{k}$ of order $k$ admits the following orthogonal decomposition:
\begin{equation}
\label{hodge_decomp}
    \mathbf{x}_{k}=\underbrace{\mathbf{B}_{k}^T\, \mathbf{x}_{k-1}}_{(a)} +\underbrace{\mathbf{B}_{k+1}\, \mathbf{x}_{k+1}}_{(b)} +\underbrace{\widetilde{\mathbf{x}}_{k}}_{(c)}  .
\end{equation}
Let us give an interpretation of the three orthogonal components in \eqref{hodge_decomp} considering edge flow signals $\mathbf{x}_{1}$ (i.e., $k=1$), as follows  \cite{barbarossa2020topological, yang2021simplicial}:

(a) Applying matrix $\mathbf{B}_{1}$ to an edge flow $\mathbf{x}_{1}$ means computing its net flow at each node, thus $\mathbf{B}_{1}$ is called a divergence operator. Its adjoint $\mathbf{B}_{1}^T $ differentiates a node signal $\mathbf{x}_{0}$ along the edges to induce an edge flow $\mathbf{B}_{1}^T\mathbf{x}_{0}$. We call $ \mathbf{B}_{1}^T\mathbf{x}_{0}$ the \textit{irrotational component} of $\mathbf{x}_{1}$ and $\textbf{im}(\mathbf{B}_{k}^T)$ the gradient space.

(b) Applying matrix $\mathbf{B}_{2}^T$ to an edge flow $\mathbf{x}_{1}$ means computing its net flow per each triangle, thus $\mathbf{B}_{2}$ is called a curl operator. Its adjoint $\mathbf{B}_{2}$ induces an edge flow $\mathbf{x}_{1}$ from a triangle signal $\mathbf{x}_{2}$. We call $\mathbf{B}_{2}\mathbf{x}_{2}$ the \textit{solenoidal component} of $\mathbf{x}_{1}$ and $\textbf{im}(\mathbf{B}_{2})$ the curl space.

(c) The remaining component $\widetilde{\mathbf{x}}_{1}$ is called the \textit{harmonic component} and  $\textbf{ker}(\mathbf{L}_{1})$ is called the harmonic space. Any edge flow $\widetilde{\mathbf{x}}_{1}$ has zero divergence and curl.

For the sake of simplicity, in the sequel we will focus on the processing of edge signals without loss of generality. Therefore, we will denote $\mathbf{x}_{1}$ with $\mathbf{x}$, $\mathbf{L}_{1}$ with $\mathbf{L}$, $\mathbf{L}_1^{(d)}$ with $\mathbf{L}^{(d)}$ and $\mathbf{L}_1^{(u)}$ with $\mathbf{L}^{(u)}$, such that $\mathbf{L} = \mathbf{L}^{(d)} + \mathbf{L}^{(u)}$. 

\textbf{Simplicial filters:} The Hodge decomposition in \eqref{hodge_decomp} suggests to separately filter the irrotational, solenoidal and harmonic components of the signal. Thus, generalizing the approach proposed in \cite{yang2021finite}, let us consider a simplicial convolutional filter design given by:
\begin{equation} \label{us_filter}
    \mathbf{H} =   \underbrace{\sum_{j = 1}^{J^{(d)}} w^{(d)}_j \big(\mathbf{L}^{(d)}\big)^j}_{\mathbf{H}^{(d)}} + \underbrace{\sum_{j = 1}^{J^{(u)}} w^{(u)}_j \big(\mathbf{L}^{(u)}\big)^j}_{\mathbf{H}^{(u)}}+\underbrace{w^{(h)}\widehat{\mathbf{P}}}_{\mathbf{H}^{(h)}}
\end{equation}
where $\mathbf{w}^{(d)} = \Big[w^{(d)}_1,...,w^{(d)}_{J^{(d)}}\Big]^T \in \mathbb{R}^{J^{(d)} \times 1}$, $\mathbf{w}^{(u)} = \Big[w^{(u)}_1,...,w^{(d)}_{J^{(u)}}\Big]^T \in \mathbb{R}^{J^{(u)}\times 1}$ and $w^{(h)} \in \mathbb{R}$ are the filter weights, $J^{(d)} \in \mathbb{N}$ is the irrotational filter order and $J^{(u)} \in \mathbb{N}$ is the solenoidal filter order. The filter in \eqref{us_filter} resembles the Hodge decomposition and it is a proper generalization to simplicial signals  of the well-known linear-shift-invariant graph filters \cite{shuman2013emerging}. In particular, the terms $\mathbf{H}^{(d)}$ and $\mathbf{H}^{(u)}$ of \eqref{us_filter} allows to independently filter the input signal based on its lower and upper simplicial neighbourhoods, thus processing its irrotational and solenoidal components, respectively. The term $\mathbf{H}^{(h)}$ extracts and scales the harmonic component of the signal, with  $\widehat{\mathbf{P}} \in \mathbb{R}^{E \times E}$ being a \textit{sparse approximated projection operator} of a true dense projection operator $\mathbf{P}$ onto the harmonic space $\textbf{ker}\big(\mathbf{L}\big)$. From \eqref{hodge_spaces} and \eqref{Laplaciank},  harmonic  signals  can  be represented  as  linear  combination  of  a  basis  of  eigenvectors  spanning  the kernel of $\mathbf{L}$. However, since there is no unique way to identify a basis for such a subspace, the approximation can be driven by ad-hoc criteria to choose a specific basis, as in \cite{sardellitti2022cell}, or just finding an approximated projector $\widehat{\mathbf{P}}$ of any of the possible bases, but with some desirable property as sparsity. In the latter case, the true dense projector should be equal to $\mathbf{P} = \widetilde{\mathbf{U}}\widetilde{\mathbf{U}}^T$, where $\widetilde{\mathbf{U}} \in \mathbb{R}^{E \times N_h}$ are eigenvectors of $\mathbf{L}$ corresponding to the zero eigenvalue of multiplicity $N_h \in \mathbb{N}$. A sparse approximation of $\mathbf{P}$ can be obtained as \cite{olfati2004consensus}:
\begin{align}\label{harmonic_filter}
    \widehat{\mathbf{P}} = \big(\mathbf{I} - \epsilon \mathbf{L}\big)^{J^{(h)}},
\end{align}
where $J^{(h)} \in \mathbb{N}$ and $0<\epsilon\leq\frac{2}{\lambda_{MAX}(\mathbf{L})}$, and such that $\displaystyle\lim_{J^{(h)} \rightarrow \infty} \; \widehat{\mathbf{P}} = \mathbf{P}$. We term $\mathbf{H}^{(h)}$ in (\ref{us_filter}) with (\ref{harmonic_filter}) as the \textit{harmonic filter}, which enjoys nice locality properties that enable distributed implementation.

\begin{figure}[t]
     \centering
              \vspace{-1.5cm}
     \begin{subfigure}[b]{0.36\textwidth}
         \centering
         \includegraphics[scale = 0.5]{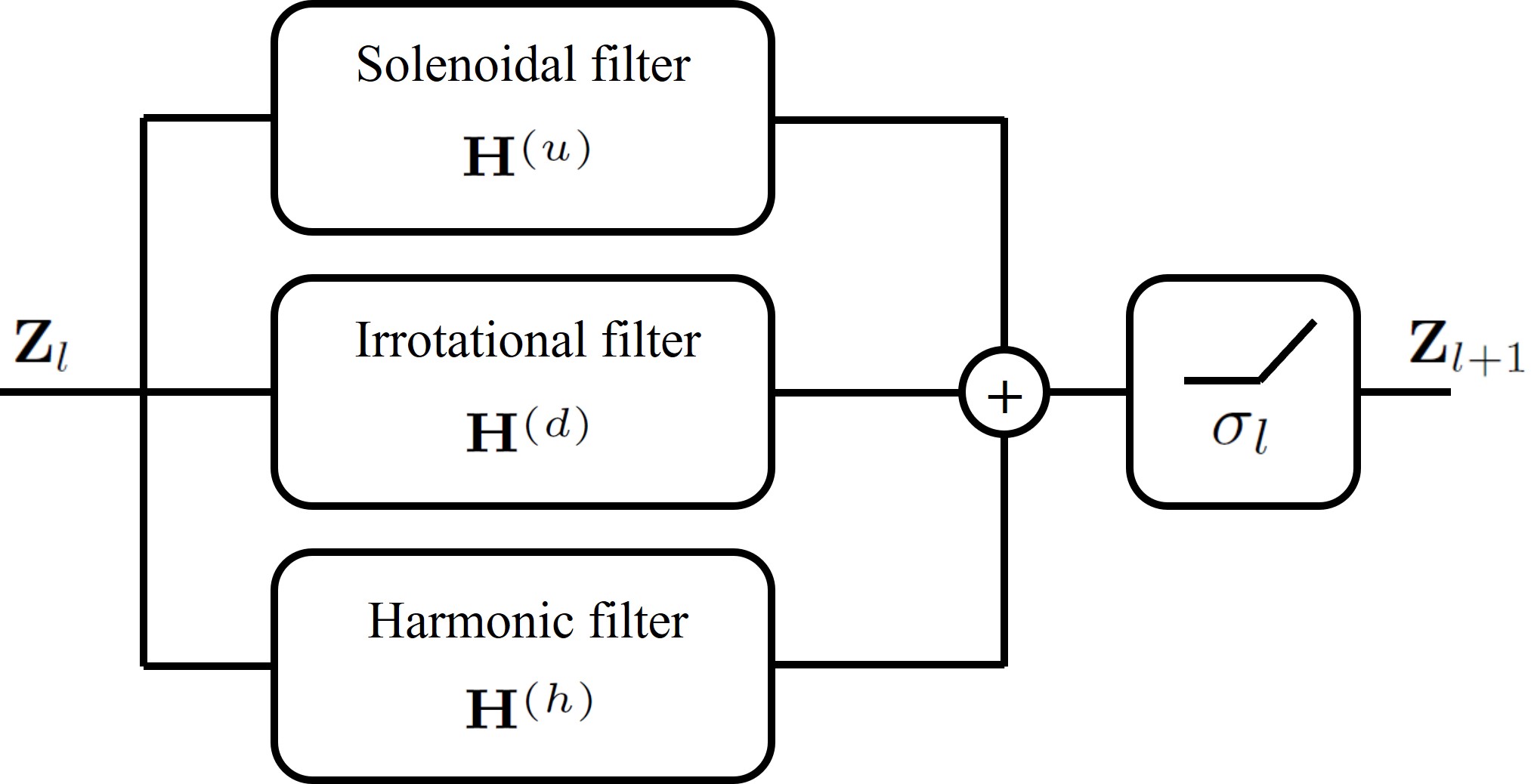}
         \caption{Simplicial attention layer}
         \label{fig:SAL}
     \end{subfigure}
     \hspace{3.5cm}
     \begin{subfigure}[b]{0.36\textwidth}
         \centering
         \includegraphics[scale = 0.57]{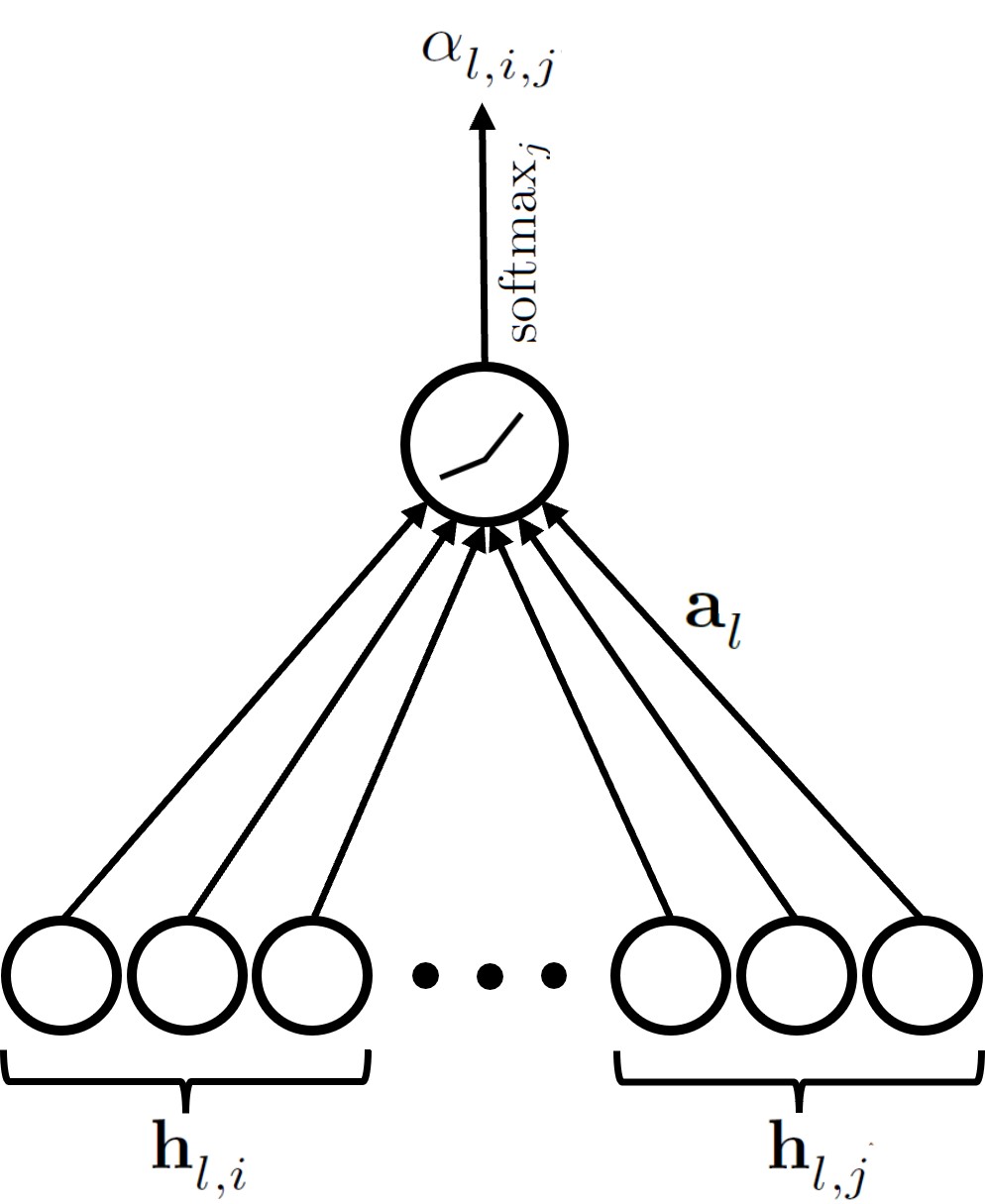}
         \caption{Simplicial attention mechanism}
         \label{fig:SAM}
     \end{subfigure}
        \caption{SAN architecture}
        \label{fig:san_figures}
\end{figure}

\textbf{Simplicial convolutional networks:} We now introduce a simplicial convolutional network (SCN) architecture, whose layers are composed of two main stages: i) simplicial filtering, and ii) point-wise non-linearity. Let us assume that $F_l$ edge signals $\{\mathbf{z}_{l,f}\}_{f=1}^{F_l}$ (also named as convolutional features in deep learning) are given as input to the $l-th$ layer of the SCN. First, each of the input signal is passed through a bank of $F_{l+1}$ filters as in \eqref{us_filter}.  
Then, the intermediate outputs  $\{\Tilde{\mathbf{z}}_{l,g,f}\}_f$ are summed to avoid exponential filter growth and, finally, a pointwise non-linearity $\sigma_l(\cdot)$ is applied. In summary, the output edge signals $\{\mathbf{z}_{l+1,g}\}_g$ of the $l-th$ layer are given by:
\begin{align} \label{SCN_layer}
    \mathbf{z}_{l+1,g} 
    = \sigma_l \Bigg(\sum_{f = 1}^{F_l}\Bigg[\sum_{p = 1}^{J_l^{(d)}} w^{(d)}_{l,p,f,g} \big(\mathbf{L}^{(d)}\big)^p + \sum_{p = 1}^{J_l^{(u)}} w^{(u)}_{l,p,f,g} \big(\mathbf{L}^{(u)}\big)^p+w^{(h)}_{l,f,g}\widehat{\mathbf{P}}_l \Bigg]\mathbf{z}_{l,f}\Bigg), 
\end{align}
for $g = 1,\ldots,F_{l+1}$.
The layer \eqref{SCN_layer} can be rewritten in a compact matrix form. Indeed, introducing the matrices $\mathbf{Z}_{l+1}=\{\mathbf{z}_{l+1,g}\}_{g=1}^{F_{l+1}}\in \mathbb{R}^{E \times F_{l+1}}$, $\mathbf{Z}_{l}=\{\mathbf{z}_{l,f}\}_{f=1}^{F_{l}}\in \mathbb{R}^{E \times F_{l}}$, $\mathbf{W}^{(d)}_{l,p} = \{w^{(d)}_{l,p,f,g}\}_{f=1,g=1}^{F_l,F_{l+1}}\in \mathbb{R}^{F_l \times F_{l+1}}$, $\mathbf{W}^{(u)}_{l,p} = \{w^{(u)}_{l,p,f,g}\}_{f=1,g=1}^{F_l,F_{l+1}}\in \mathbb{R}^{F_l \times F_{l+1}}$, and
$\mathbf{W}^{(h)}_{l} = \{w^{(h)}_{l,f,g}\}_{f=1,g=1}^{F_l,F_{l+1}}\in \mathbb{R}^{F_l \times F_{l+1}}$, the layer \eqref{SCN_layer} can be equivalently recast as: 
\begin{equation} \label{SCN_layer_matrix}
\mathbf{Z}_{l+1} = \sigma_l \Bigg( \sum_{p = 1}^{J_l^{(d)}}(\mathbf{L}^{(d)})^p\mathbf{Z}_l\mathbf{W}^{(d)}_{l,p} +  \sum_{p = 1}^{J^{(u)}_l} (\mathbf{L}^{(u)})^p\mathbf{Z}_l\mathbf{W}^{(u)}_{l,p}  + \widehat{\mathbf{P}}_l\mathbf{Z}_l\mathbf{W}^{(h)}_l \Bigg).
\end{equation}
The filters weights $\big\{\mathbf{W}^{(d)}_{l,p}\big\}_{p=1}^{J^{(d)}_l}$, $\big\{\mathbf{W}^{(u)}_{l,p}\big\}_{p=1}^{J^{(u)}_l}$ and $\mathbf{W}^{(h)}_{l}$ are learnable parameters, while the order $J^{(d)}_l$ and $J^{(u)}_l$ of the filters, the number $F_{l+1}$ of output signals, and the non-linearity $\sigma_l(\cdot)$ are hyperparameters to be chosen at each layer. Therefore, a SCN of depth $L$ with input data $\mathbf{X} \in \mathbb{R}^{E \times F_0}$ is built as the stack of $L$ layers defined as in \eqref{SCN_layer_matrix}, where $\mathbf{Z}_0 = \mathbf{X}$. Based on the learning task, an additional multi layer perceptron (MLP) can be inserted after the last layer. Please notice that the layer in \eqref{SCN_layer_matrix} can be clearly interpreted as the aggregation of three filtering branches, one for irrotational components (based on lower neighborhoods), one for solenoidal components (based on upper neighborhoods), and one for the harmonic component, as illustrated in Fig. \ref{fig:san_figures}(a).

\section{Simplicial Attention  Neural  Networks}
In this section, we  extend the SCN layer in \eqref{SCN_layer_matrix} introducing our SAN architecture. We also analyze SAN's computational complexity, and its relations with other simplicial neural architectures.

\subsection{Simplicial Attention Layer}

The proposed attentional mechanism follows the approach from \cite{velivckovic2017graph,bahdanau2014neural}, which is then generalized to handle high-order simplicial complexes. The core idea is to learn the  Laplacian matrices $\mathbf{L}^{(u)}$ and $\mathbf{L}^{(d)}$ in (\ref{SCN_layer_matrix}), in order to optimally combine data over upper and lower neighborhoods in a totally data-driven fashion. To this aim, we will denote the Laplacians associated to the $l$-th layer with $\mathbf{L}^{(u)}_l$ and $\mathbf{L}^{(d)}_l$, and we will refer to them as \textit{attentional Laplacians}. The input to our layer is $\mathbf{Z}_l$, and the first step is applying a collection of shared learnable linear transformations parametrized by the filter weights $\big\{\mathbf{W}^{(d)}_{l,p}\big\}_{p=1}^{J_l^{(d)}}$ and $\big\{\mathbf{W}^{(u)}_{l,p}\big\}_{p=1}^{J_l^{(u)}}$ over lower and upper branches, respectively. Let $\mathbf{h}_{l,p,i}^{(u)}=[\mathbf{Z}_l\big]_i\mathbf{W}^{(u)}_{l,p}\in \mathbb{R}^{F_{l+1}}$, for $p=1,\ldots,J_l^{(u)}$, and $\mathbf{h}_{l,p,i}^{(d)}=[\mathbf{Z}_l\big]_i\mathbf{W}^{(d)}_{l,p}\in \mathbb{R}^{F_{l+1}}$, for $p=1,\ldots,J_l^{(d)}$, be the linear transformation of the input convolutional features. Also, let us collect all the transformed features into the vectors
$$\mathbf{h}_{l,i}^{(u)} =  \{\mathbf{h}_{l,p,i}^{(u)}\}_{p=1}^{J_l^{(u)}}\in \mathbb{R}^{F_{l+1}J^{(u)}_l}, \qquad\mathbf{h}_{l,i}^{(d)} =  \{\mathbf{h}_{l,p,i}^{(d)}\}_{p=1}^{J_l^{(d)}}\in \mathbb{R}^{F_{l+1}J^{(d)}_l}.$$ 
Then, we perform topology-aware self-attention on the edges (or higher-order simplices). An upper attentional mechanism $a^{(u)}_l :\mathbb{R}^{F_{l+1}J^{(u)}_l} \hspace{-.1cm} \times  \mathbb{R}^{F_{l+1}J^{(u)}_l} \rightarrow \mathbb{R}$ and a lower attentional  mechanism $a^{(d)}_l : \mathbb{R}^{F_{l+1}{J^{(d)}_l}} \hspace{-.1cm} \times  \mathbb{R}^{F_{l+1}J^{(d)}_l}  \rightarrow \mathbb{R}$ compute the attention coefficients:
\begin{align}\label{attention_coefficients}
    &  e_{l,i,j}^{(u)} = a^{(u)}_l\hspace{-.1cm}\left(\mathbf{h}_{l,i}^{(u)},\mathbf{h}_{l,j}^{(u)} \right) \qquad \hbox{for $j \in \mathcal{N}_i^{(u)}$},  \\
    &e_{l,i,j}^{(d)} = a^{(d)}_l\hspace{-.1cm}\left(\mathbf{h}_{l,i}^{(d)},\mathbf{h}_{l,j}^{(d)} \right) \qquad \hbox{for $j \in \mathcal{N}_i^{(d)}$}, 
\end{align}
which represent the importance of edge $j$’s features to edge $i$ over upper (i.e., $\mathcal{N}_i^{(u)}$) and lower (i.e., $\mathcal{N}_i^{(d)}$) neighborhoods, respectively. To make coefficients easily comparable across different edges, we normalize them across all choices of j using the softmax function:
\begin{align}\label{softmax}
    &  \alpha_{l,i,j}^{(u)} = {\rm softmax}_j (e_{l,i,j}^{(u)})=\frac{{\rm exp}(e_{l,i,j}^{(u)})}{\sum_{k\in \mathcal{N}_i^{(u)}}{\rm exp}(e_{l,i,k}^{(u)})} \qquad \hbox{for $j \in \mathcal{N}_i^{(u)}$},  \\
    &\alpha_{l,i,j}^{(d)} = {\rm softmax}_j (e_{l,i,j}^{(d)})=\frac{{\rm exp}(e_{l,i,j}^{(d)})}{\sum_{k\in \mathcal{N}_i^{(d)}}{\rm exp}(e_{l,i,k}^{(d)})} \qquad \hbox{for $j \in \mathcal{N}_i^{(d)}$}.
\end{align}
In this paper, we exploit single-layer feedforward neural networks as possible attention mechanism, which is parametrized by the weight vectors $\mathbf{a}^{(u)}_l\in \mathbb{R}^{2 J^{(u)} F_{l+1}}$ and $\mathbf{a}^{(d)}_l\in \mathbb{R}^{2 J^{(d)} F_{l+1}}$ for the upper and lower components, respectively. Then, exploiting a LeakyReLU nonlinearity,  the attention coefficients (illustrated by Fig. \ref{fig:san_figures} (b)) can be finally expressed as: 
\begin{align}
    &  \alpha_{l,i,j}^{(u)} =\frac{{\rm exp}\left(\textrm{LeakyReLU}\left( \left[\mathbf{h}_{l,i}^{(u)} \| \mathbf{h}_{l,j}^{(u)}\right]^T \mathbf{a}^{(u)}_l \right)\right)}{\sum_{k\in \mathcal{N}_i^{(u)}}{\rm exp}\left(\textrm{LeakyReLU}\left( \left[\mathbf{h}_{l,i}^{(u)} \| \mathbf{h}_{l,k}^{(u)}\right]^T \mathbf{a}^{(u)}_l \right)    \right)}  \qquad \hbox{for $j \in \mathcal{N}_i^{(u)}$},\label{attentionNN1}\\
    &\alpha_{l,i,j}^{(d)} =\frac{{\rm exp}\left(\textrm{LeakyReLU}\left( \left[\mathbf{h}_{l,i}^{(d)} \| \mathbf{h}_{l,j}^{(d)}\right]^T \mathbf{a}^{(d)}_l \right)\right)}{\sum_{k\in \mathcal{N}_i^{(d)}}{\rm exp}\left(\textrm{LeakyReLU}\left( \left[\mathbf{h}_{l,i}^{(d)} \| \mathbf{h}_{l,k}^{(d)}\right]^T \mathbf{a}^{(d)}_l \right)    \right)}\qquad \hbox{for $j \in \mathcal{N}_i^{(d)}$}.\label{attentionNN2}
\end{align}
where $\|$ denotes the vector concatentation operator. Finally, from (\ref{SCN_layer_matrix}), the SAN layer can be cast as
\begin{equation} \label{SAN_layer_matrix}
\mathbf{Z}_{l+1} = \sigma_l \Bigg(\underbrace{\sum_{p = 1}^{J^{(d)}_l}\big(\mathbf{L}^{(d)}_l\big)^p\mathbf{Z}_l\mathbf{W}^{(d)}_{l,p} +  \sum_{p = 1}^{J^{(u)}_l} \big(\mathbf{L}^{(u)}_l\big)^p\mathbf{Z}_l\mathbf{W}^{(u)}_{l,p}  + \widehat{\mathbf{P}}_l\mathbf{Z}_l\mathbf{W}^{(h,l)}}_{\widetilde{\mathbf{Z}}_{l+1}} \Bigg),
\end{equation}
where the coefficients of the upper and lower attentional Laplacians $\mathbf{L}^{(u)}_l$ and $\mathbf{L}^{(d)}_l$ are obtained as in \eqref{attentionNN1}-\eqref{attentionNN2}, respectively. The filters weights $\big\{\mathbf{W}^{(d)}_{l,p}\big\}_p$, $\big\{\mathbf{W}^{(u)}_{l,p}\big\}_p$, $\mathbf{W}^{(h)}_l$ and the attention mechanism parameters $\mathbf{a}^{(u)}_l $ and $\mathbf{a}^{(d)}_l $ are learnable parameters, while the order $J^{(d)}_l$ and $J^{(u)}_l$ of the filters, the number $F_{l+1}$ of output signals, and the non-linearity $\sigma_l(\cdot)$ 
are hyperparameters to be chosen at each layer. Therefore, a SAN of depth $L$ with input data $\mathbf{X} \in \mathbb{R}^{E \times F_0}$ is built as the stack of $L$ layers defined as in \eqref{SCN_layer_matrix}, where $\mathbf{Z}_0 = \mathbf{X}$. Based on the learning task, an additional multi-layer perceptron (MLP) can be inserted after the last layer. Clearly, the same principled attention mechanism can be applied to process data defined over higher-order simplices, e.g., triangles, tetrahedra, and so on. Finally, notice that $K$-heads (multi-head) attention can be straightforwardly implemented by horizontally concatenating or averaging $K$ outputs  $\widetilde{\mathbf{Z}}^k_{l+1}$ as in \eqref{SAN_layer_matrix}. In such cases, the final layer output writes as $\mathbf{Z}_{l+1} =   \left\{\sigma_l \Big(\widetilde{\mathbf{Z}}^k_{l+1}\Big)\right\}_{k=1}^K  \in \mathbb{R}^{E \times KF_{l+1}}$ and $\mathbf{Z}_{l+1} =  \sigma_l \left( \frac{1}{K}\sum_{k=1}^K\widetilde{\mathbf{Z}}^k_{l+1} \right)  \in \mathbb{R}^{E \times F_{l+1}}$, respectively.

\subsection{Complexity Analysis}\label{sec:cmpl-an}
The total number of learnable parameters of a SAN layer with single-head attention as in \eqref{SAN_layer_matrix} is  $2J^{(d)}_lF_{l+1}+2J^{(u)}_lF_{l+1}+(J^{(u)}_l+J^{(u)}_l)F_{l}F_{l+1}+F_{l}F_{l+1}$. From a computational point of view, the SAN architecture is highly efficient and the complexity comes from two main sources: 
\begin{enumerate}
    \item Convolutional filtering is a local operation within the simplicial neighbourhoods and can be computed recursively. Therefore, the overall complexity  of the filtering stage is in the order of $\mathcal{O}(U(J^{(u)}_l+J^{(d)}_l)F_{l}F_{l+1})$ per each edge, where $U$ is maximum number of neighbours;
    \item The computation of the attention coefficients in (\ref{attentionNN1})-(\ref{attentionNN2}) is also a local operation within the simplicial neighbourhoods. Therefore, the overall complexity of the attentional mechanism is in the order of $\mathcal{O}(U(2(J^{(d)}_l+J^{(u)}_l)F_{l}F_{l+1}+ 2(J^{(d)}_l+J^{(u)}_l)F_{l+1}))$ for each edge.
\end{enumerate}
In the multi-head case, the previous derivations are multiplied by a factor $K$. Please notice that the local operations can be further parallelized also across filtering branches. The previous complexity analysis assume classical dense-dense benchmark algorithms for matrix and vectors multiplication. Efficiency can  be further improved using sparse-dense or sparse-sparse algorithms in the case of sparse topologies. Finally, principled topology-dependent implementations can be exploited. Although generally improving efficiency and scalability, parallelization across all the edges and branches may involve redundant computation, as the neighborhoods will often overlap in the topology of interest.

\vspace{-.1cm}
\subsection{Comparisons with related works}

The proposed architecture in \eqref{SAN_layer_matrix} generalizes most of the simplicial neural network architectures available in literature and, clearly, also the graph attention network introduced in \cite{velivckovic2017graph}. In particular, the architecture proposed in \cite{ebli2020simplicial} can be obtained from \eqref{SAN_layer_matrix} removing the attentional mechanism, the harmonic filtering and setting $\{\mathbf{W}^{(d)}_{l,p}\}_p =\{\mathbf{W}^{(u)}_{l,p}\}_p$ and $J^{(u)}_l=J^{(d)}_l=1$. The architecture in \cite{yang2021simplicial} can be obtained from  \eqref{SAN_layer_matrix} removing the attentional mechanism and setting $J^{(h)}_l=0$. The architectures in \cite{bunch2020simplicial} and \cite{roddenberry2019hodgenet} can be obtained removing the attentional mechanism, and setting  $J^{(u)}_l=J^{(d)}_l=1$ and $J^{(h)}_l=0$. The SAT architecture in \cite{anonymous2022SAT} with sum as aggregation function can be obtained from \eqref{SAN_layer_matrix} removing the harmonic filtering, setting  $J^{(u)}_l=J^{(d)}_l=1$, \textcolor{black}{and considering a single attention mechanism over upper and lower neighborhoods, i.e., $a_{l}^{(u)}=a_{l}^{(d)}=a_{l}$, for all $l$.} Finally,  GAT in \cite{velivckovic2017graph}  can be obtained in the case of $0$-simplex signals, removing the harmonic filtering and setting $J^{(u)}_l=1$ ($J^{(u)}_l$ does not exist for node signals).

\vspace{-.2cm}
\section{Experimental Results}
\vspace{-.1cm}

In this section, we assess the performance of SANs on two challenging tasks: Trajectory prediction \cite{bodnar2021weisfeiler} (inductive learning), and missing data imputation in citation complexes \cite{ebli2020simplicial,yang2021simplicial} (transductive learning), illustrating how the proposed architecture improves current state-of-the-art results \footnote{Our SAN implementation \& datasets are available at \url{https://github.com/lrnzgiusti/Simplicial-Attention-Networks}} .

\begin{figure}[t]

     \centering
     
     \begin{subfigure}[b]{0.4\textwidth}
         \centering
         \includegraphics[scale = 0.48]{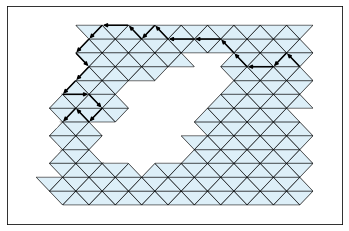}
         \caption{Ocean Drifters Map}
         \label{fig:ocean}
     \end{subfigure}
     \hspace{1.5cm}
     \begin{subfigure}[b]{0.4\textwidth}
         \centering
         \includegraphics[scale = 0.15]{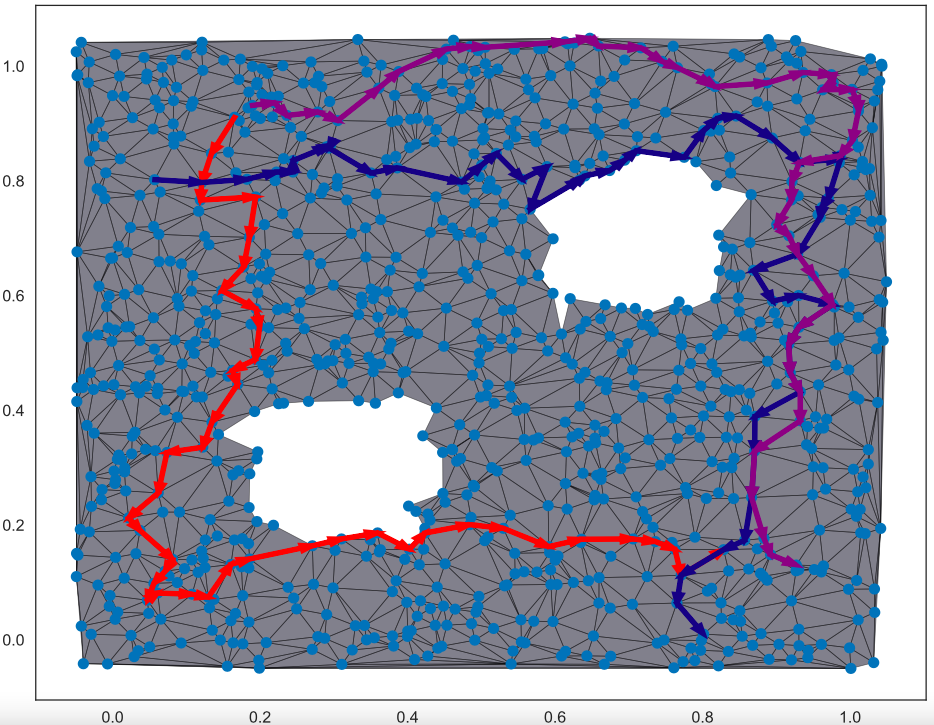}
         \caption{Synthetic Flow Map}
         \label{fig:synth}
     \end{subfigure}
        \caption{Setup of trajectory prediction task. (a) Ocean drifters flow around the Madagascar island. (b) Synthetic flows over a simplicial complex.}
        \label{fig:maps}
\end{figure}

\begin{table}[t]
\caption{Trajectory classification test accuracy.}
\centering
\label{tab:trajectory}
\begin{adjustbox}{max width=\textwidth}
\begin{tabular}{|cccc|}
\hline

\begin{tabular}[c]{@{}c@{}}Architecture\end{tabular} & 
\begin{tabular}[c]{@{}c@{}}Activation   \end{tabular}           & 
\begin{tabular}[c]{@{}c@{}}Synthetic Flow (\%)\end{tabular}                                       & 
\multicolumn{1}{c|}{\begin{tabular}[c]{@{}c@{}}Ocean Drifters (\%)\end{tabular}}            \\ \hline

MPSN \; \cite{bodnar2021weisfeiler}                                                 &
\begin{tabular}[c]{@{}c@{}}Id\\ ReLU \\ Tanh \end{tabular} & 
\begin{tabular}[c]{@{}c@{}}82.6 $\pm$ 3.0\\ 50.0  $\pm$ 0.0\\ 95.2 $\pm$ 1.8 \end{tabular} & 
\multicolumn{1}{c|}{\begin{tabular}[c]{@{}c@{}}73.0 $\pm$ 2.7\\ 46.5 $\pm$ 5.7\\ 72.5 $\pm$ 0.0\end{tabular}}\\ \hline

SCNN \; \cite{yang2021simplicial}                                              & 
\begin{tabular}[c]{@{}c@{}}Id \\ ReLU\\ Tanh\end{tabular} & 
\begin{tabular}[c]{@{}c@{}}66.5 $\pm$ 0.16\\ 100 $\pm$ 0.0\\ 67.2  $\pm$ 0.16 \end{tabular} & 
\begin{tabular}[c]{@{}l@{}}98.1 $\pm$ 0.01\\ 97.0 $\pm$ 0.01\\ 97.0  $\pm$ 0.16  \end{tabular}           \\ \hline

SAT \; \cite{anonymous2022SAT}                                            & 
\begin{tabular}[c]{@{}c@{}}Id\\ ReLU \\ Tanh \end{tabular} & 
\begin{tabular}[c]{@{}c@{}}99.7 $\pm$ 0.0\\ 100 $\pm$ 0.0\\ 100 $\pm$ 0.0\end{tabular} & 
\begin{tabular}[c]{@{}l@{}}97.0 $\pm$ 0.01\\ 95.0 $\pm$ 0.00\\ 95.0 $\pm$ 0.01 \end{tabular}                      \\ \hline

SAN $\left(J^{(h)} = 0\right)$                                               & 
\begin{tabular}[c]{@{}c@{}}Id \\ ReLU\\ Tanh\end{tabular} & 
\begin{tabular}[c]{@{}c@{}}100 $\pm$ 0.0\\ 100 $\pm$ 0.0\\ 100 $\pm$ 0.0 \end{tabular} & 
\begin{tabular}[c]{@{}l@{}}97.1 $\pm$ 0.02\\ 97.0 $\pm$ 0.2\\ 97.5 $\pm$ 0.02\end{tabular}          \\ \hline

SAN                                                & 
\begin{tabular}[c]{@{}c@{}}Id \\ ReLU\\ Tanh \end{tabular} & 
\begin{tabular}[c]{@{}c@{}}100 $\pm$ 0.0\\ 100 $\pm$ 0.0\\ \tb{100} $\pm$ \tb{0.0}\end{tabular} & 
\begin{tabular}[c]{@{}l@{}}\tb{99.0} $\pm$ \tb{0.01}\\ 98.5 $\pm$ 0.01\\ 98.5 $\pm$ 0.01\end{tabular}             \\ \hline
\end{tabular}
\end{adjustbox}
\end{table}

\vspace{-.2cm}
\subsection{Trajectory Prediction}
\vspace{-.1cm}

Trajectory prediction tasks have been adopted to solve many problems in location-based services, e.g. route recommendation \cite{zheng2014modeling}, or inferring the missing portions of a given trajectory \cite{wu2016probabilistic}. 
Inspired by  \cite{schaub2020random}, the works in \cite{roddenberry2021principled,bodnar2021weisfeiler} exploited simplicial neural networks to tackle trajectory prediction problems. In the sequel, for comparison purposes, we follow the same simulation setup of \cite{bodnar2021weisfeiler}.\smallskip\\
\textbf{Synthetic Flow:} We first test our architecture on the synthetic flow dataset from \cite{bodnar2021weisfeiler}. The simplicial complex is generated by sampling $400$ points uniformly at random in the unit square, and then a Delaunay triangulation is applied to obtain the domain of the trajectories. The set of trajectories is generated on the simplicial complex shown in figure \ref{fig:synth}: Each trajectory starts from the top left corner and go through the whole map until the bottom right corner, passing close to either the bottom-left hole or the top-right hole. Thus, the learning task is to identify which of the two holes is the closest one on the path. The dataset is made of $1000$ training examples and $200$ test data. \smallskip\\ \textbf{Ocean Drifters:} We also consider a real-world dataset including ocean drifter tracks near Madagascar from 2011 to 2018 \cite{schaub2020random}. The map surface is discretized into a simplicial complex with a hole in the centre, which represents the presence of the island. The discretization process is done by tiling the map into a regular hexagonal grid. Each hexagon represents a $0$-simplex (vertex), and if there is a nonzero net flow from one hexagon to its surrounding neighbors, a $1$-simplex (edge) is placed between them. All the 3-cliques of the $1$-simplex are considered to be $2$-simplex (triangles) of the simplicial complex shown in figure \ref{fig:ocean}. Thus, following the experimental setup of \cite{bodnar2021weisfeiler}, the learning task is to distinguish between the clockwise and counter-clockwise motions of flows around the island. The dataset is composed of $160$ training trajectories and $40$ test trajectories. The flows belonging to  each trajectory of the test set use random orientations. \smallskip\\
Both experiments are inductive learning problems. In both cases, we exploited a single layer SAN architecture with $4$ output features, and upper and lower filter lengths $J^{(u)} = J^{(d)} = 3$. To perform the classification task, we used an MLP as readout layer with softmax non-linearity. Regarding the harmonic projection, we set the parameters of \eqref{harmonic_filter} as $\epsilon=0.9$ and $J^{(h)}=5$. The network is trained via ADAM optimizer \cite{kingma2017adam} and cross-entropy loss,  with initial learning rate set to $0.01$, a step reduction of $0.77$  and a patience of $10$ epochs. To avoid overfitting, we exploited $l_2$ regularization with $\lambda_{l_2} = 0.003$ and dropout with $p=0.6$. In Table \ref{tab:trajectory} we compare the accuracy of the proposed SAN architecture averaged over $5$ different seeds; for each seed, the networks is trained with an early stopping criteria with a patience of $100$ epochs, with and without (i.e., $J^{(h)}=0$) harmonic filtering. We compare it against the MPSN architecture from \cite{bodnar2021weisfeiler}, the SCN architecture (same hyperparameters as ours) from \cite{yang2021simplicial}, and the  architecture from \cite{anonymous2022SAT}, referred as SAT, (same hyperparameters as ours, but with $J^{(d)}=J^{(u)}=1$ and no harmonic projection by architecture definition), with sum as aggregation function, for different activation functions of the simplicial layer. Both SAT and SAN exploit single-head attention.  Unfortunately, we were unable to find any implementation or reproducible experimental set up for the SAT architecture in \cite{anonymous2022SAT}, thus we have chosen the fairest architecture in terms of complexity and structure, being it an instance  both of a SAN and a SAT. For the MPSN architecture, we reuse the metrics already reported in  \cite{bodnar2021weisfeiler}. As the reader can notice from Table \ref{tab:trajectory}, the proposed SAN architecture achieves the best results among all the competitors in both the synthetic and real-world datasets. In the synthetic example, our architecture achieves 100\% of accuracy independently on the used non-linearity. 

\vspace{-.2cm}
\subsection{Citation Complex Imputation}

Missing data imputation (MDI) is a learning task that consists in estimating missing values in a dataset.  
GNN can be used to tackle this task as in \cite{spinelli2019ginn}, but recently the works in \cite{ebli2020simplicial, yang2021simplicial} have handled the MDI problem using simplicial complexes. We followed the experimental settings of \cite{ebli2020simplicial}, estimating the number of citation of a collaboration between $k+1$ authors over a co-authorship complex domain. This is as a transductive learning task, where the labels of the $k$-simplex are the number of citation of the $k+1$ authors. We employ a four-layers SAN architecture, with $256$ output feature for the first three layers, and a filter length over upper and lower neighborhoods $J^{(u)}_l = J^{(d)}_l = 2$, for all $l$. The final layer computes a single output feature that will be used as estimate of the $k$-simplex' labels. We employ ReLu non-linearity per each layer. We have found that setting $J^{(h)}_l=0$ for all $l$ (i.e., no harmonic projection) is the best option for this task. In this case, the harmonic projection stage turns into a simple skip connection \cite{he2016deep}. To train the network, we used a Xavier Uniform initialization with a gain of $\sqrt{2}$,  ADAM optimizer  with $0.1$ as initial learning rate equipped with a step reduction on plateaus with a patience of $100$ epochs, and masked $\ell_1$ loss with an early stopping criteria with patience of $500$ epochs. Accuracy is computed considering a citation value correct if its estimate is within $\pm5\%$ of the true value. In Table \ref{tab:citation}, we illustrate the mean performance and the standard deviation of our architecture over $10$ different \textcolor{black}{masks for missing data}. Comparing the results with SNN in \cite{ebli2020simplicial}, SCNN in \cite{yang2021simplicial}, and SAT (same considerations as previous experiment) in \cite{anonymous2022SAT}, for different simplex orders and percentages of missing data. Both SAT and SAN exploit single-head attention. Moreover, to fairly evaluate the benefits of the attention mechanism, we also compare the proposed SAN architecture with a SCNN \cite{yang2021simplicial} of the same size and hyperparameters, denoted by SCNN(ours). As we can notice from Table \ref{tab:citation}, SAN achieves the best performance per each order and percentage of missing data, with huge gains as the order and the percentage grow, illustrating the importance of incorporating principled attention mechanisms in simplicial neural architectures. \textcolor{black}{Indeed, also SAT performs very poorly, due to its upper-lower shared attention mechanism and its impossibility of processing the harmonic component (or exploiting skip connections).}

\begin{table}[t]
\caption{Missing Data Imputation test accuracy}
\centering
\label{tab:citation}
\begin{adjustbox}{max width=0.9\textwidth}
\begin{tabular}{|cccccccc|}
\hline

\begin{tabular}[c]{@{}c@{}}\%Miss/Order\\ $N_k$\end{tabular} & 
\begin{tabular}[c]{@{}c@{}}Method   \end{tabular}           & 
\begin{tabular}[c]{@{}c@{}}0\\ 352\end{tabular}                                       & 
\begin{tabular}[c]{@{}c@{}}1\\ 1474\end{tabular}                                      & 
\begin{tabular}[c]{@{}c@{}}2\\ 3285\end{tabular}                                      & 
\begin{tabular}[c]{@{}c@{}}3\\ 5019\end{tabular}                                      & 
\begin{tabular}[c]{@{}c@{}}4\\ 5559\end{tabular}                                      & 
\multicolumn{1}{c|}{\begin{tabular}[c]{@{}c@{}}5\\ 4547\end{tabular}}            \\ \hline

10\%                                                 &
\begin{tabular}[c]{@{}c@{}}SNN \; \cite{ebli2020simplicial}\\ SCNN\;\cite{yang2021simplicial}\\SCNN (ours)\\SAT \\SAN\end{tabular} & 
\begin{tabular}[c]{@{}c@{}}91 $\pm$ 0.3\\ 91  $\pm$ 0.4\\ 90 $\pm$ 0.3\\ 18 $\pm$ 0.0 \\ \tb{91} $\pm$ 0.4 \end{tabular} & 
\begin{tabular}[c]{@{}c@{}}91 $\pm$ 0.2\\ 91  $\pm$ 0.2\\ 91 $\pm$ 0.3\\ 31 $\pm$ 0.0 \\ \tb{95} $\pm$ \tb{1.9}\end{tabular} & 
\begin{tabular}[c]{@{}c@{}}91 $\pm$ 0.2\\ 91  $\pm$ 0.2\\ 91 $\pm$ 0.3\\ 28 $\pm$ 0.1 \\ \tb{95} $\pm$ \tb{1.9}\end{tabular} & 
\begin{tabular}[c]{@{}c@{}}91 $\pm$ 0.2\\ 91  $\pm$ 0.2\\ 93 $\pm$ 0.2\\ 34 $\pm$ 0.1 \\ \tb{97} $\pm$ \tb{1.6}\end{tabular} & 
\begin{tabular}[c]{@{}c@{}}91 $\pm$ 0.2\\ 91  $\pm$ 0.2\\ 92 $\pm$ 0.2\\ 53 $\pm$ 0.1 \\ \tb{98} $\pm$ \tb{0.9}\end{tabular} & 
\multicolumn{1}{c|}{\begin{tabular}[c]{@{}c@{}}90 $\pm$ 0.4\\ 91 $\pm$ 0.2\\ 94 $\pm$ 0.1\\ 55 $\pm$ 0.1z \\ \tb{98} $\pm$ \tb{0.7} \end{tabular}}\\ \hline

20\%                                                 & 
\begin{tabular}[c]{@{}c@{}}SNN \cite{ebli2020simplicial}\\ SCNN \cite{yang2021simplicial}\\SCNN (ours)\\SAT \\ SAN\end{tabular} & 
\begin{tabular}[c]{@{}c@{}}81 $\pm$ 0.6\\ 81 $\pm$ 0.7\\ 81 $\pm$ 0.6\\ 18 $\pm$ 0.0   \\ \tb{82} $\pm$ \tb{0.8} \end{tabular} & 
\begin{tabular}[c]{@{}c@{}}82 $\pm$ 0.3\\ 82 $\pm$ 0.3\\ 83 $\pm$ 0.7\\ 30 $\pm$ 0.0   \\ \tb{91} $\pm$ \tb{2.4} \end{tabular} & 
\begin{tabular}[c]{@{}c@{}}81 $\pm$ 0.6\\ 81 $\pm$ 0.7\\ 81 $\pm$ 0.6\\ 29 $\pm$ 0.1   \\ \tb{82} $\pm$ \tb{0.8} \end{tabular} & 
\begin{tabular}[c]{@{}c@{}}82 $\pm$ 0.3\\ 82 $\pm$ 0.3\\ 88 $\pm$ 0.4\\ 35 $\pm$ 0.1   \\ \tb{96} $\pm$ \tb{0.4} \end{tabular} & 
\begin{tabular}[c]{@{}c@{}}81 $\pm$ 0.6\\ 81 $\pm$ 0.7\\ 86 $\pm$ 0.7 \\ 50 $\pm$ 0.1  \\ \tb{96} $\pm$ \tb{1.3} \end{tabular} & 
\begin{tabular}[c]{@{}l@{}}82 $\pm$ 0.5\\ 83 $\pm$ 0.3\\ 89 $\pm$ 0.6 \\ 58 $\pm$ 0.1  \\ \tb{97} $\pm$ \tb{0.9} \end{tabular}                      \\ \hline

30\%                                                 & 
\begin{tabular}[c]{@{}c@{}}SNN \cite{ebli2020simplicial}\\ SCNN \cite{yang2021simplicial}\\SCNN (ours)\\SAT \\ SAN\end{tabular} & 
\begin{tabular}[c]{@{}c@{}}72 $\pm$ 0.6\\ 72 $\pm$ 0.5\\ 72  $\pm$ 0.6 \\ 19 $\pm$ 0.0 \\ \tb{75} $\pm$ \tb{2.1}\end{tabular} & 
\begin{tabular}[c]{@{}c@{}}73 $\pm$ 0.4\\ 73 $\pm$ 0.4\\ 76  $\pm$ 0.6 \\ 33 $\pm$ 0.1 \\ \tb{89} $\pm$ \tb{2.1}\end{tabular} & 
\begin{tabular}[c]{@{}c@{}}81 $\pm$ 0.6\\ 81 $\pm$ 0.7\\ 81 $\pm$ 0.6 \\ 25 $\pm$ 0.1 \\ \tb{82} $\pm$ \tb{0.8} \end{tabular} & 
\begin{tabular}[c]{@{}c@{}}82 $\pm$ 0.3\\ 82 $\pm$ 0.3\\ 82 $\pm$ 1.2 \\ 33 $\pm$ 0.0 \\ \tb{94} $\pm$ \tb{0.4} \end{tabular} & 
\begin{tabular}[c]{@{}c@{}}81 $\pm$ 0.6\\ 81 $\pm$ 0.7\\ 80 $\pm$ 0.7 \\ 47 $\pm$ 0.1 \\ \tb{95} $\pm$ \tb{0.5} \end{tabular} & 
\begin{tabular}[c]{@{}l@{}}73 $\pm$ 0.5\\ 74 $\pm$ 0.3\\ 86  $\pm$ 0.8  \\ 53 $\pm$ 0.1 \\ \tb{96} $\pm$ \tb{0.5}\end{tabular}           \\ \hline

40\%                                                 & 
\begin{tabular}[c]{@{}c@{}}SNN \cite{ebli2020simplicial}\\ SCNN \cite{yang2021simplicial}\\SCNN (ours)\\SAT \\ SAN\end{tabular} & 
\begin{tabular}[c]{@{}c@{}}63 $\pm$ 0.7\\ 63 $\pm$ 0.6\\ 63 $\pm$ 0.7 \\ 20 $\pm$ 0.0 \\ \tb{67} $\pm$ \tb{1.9}\end{tabular} & 
\begin{tabular}[c]{@{}c@{}}64 $\pm$ 0.3\\ 64 $\pm$ 0.3\\ 67 $\pm$ 1.1 \\ 29 $\pm$ 0.0 \\ \tb{85} $\pm$ \tb{2.8}\end{tabular} & 
\begin{tabular}[c]{@{}c@{}}81 $\pm$ 0.6\\ 81 $\pm$ 0.7\\ 81 $\pm$ 0.6 \\ 22 $\pm$ 0.0 \\ \tb{82} $\pm$ \tb{0.8} \end{tabular} & 
\begin{tabular}[c]{@{}c@{}}82 $\pm$ 0.3\\ 82 $\pm$ 0.3\\ 79 $\pm$ 1.0 \\ 43 $\pm$ 0.1 \\ \tb{91} $\pm$ \tb{0.9} \end{tabular} & 
\begin{tabular}[c]{@{}c@{}}81 $\pm$ 0.6\\ 81 $\pm$ 0.7\\ 74 $\pm$ 1.1 \\ 51 $\pm$ 0.1  \\ \tb{93} $\pm$ \tb{1.1} \end{tabular} & 
\begin{tabular}[c]{@{}l@{}}65 $\pm$ 0.3\\ 65 $\pm$ 0.2\\ 83 $\pm$ 0.9 \\ 50 $\pm$ 0.1 \\ \tb{95} $\pm$ \tb{1.6}\end{tabular}          \\ \hline

50\%                                                 & 
\begin{tabular}[c]{@{}c@{}}SNN \cite{ebli2020simplicial}\\ SCNN \cite{yang2021simplicial}\\SCNN (ours)\\SAT \\ SAN\end{tabular} & 
\begin{tabular}[c]{@{}c@{}}54 $\pm$ 0.7\\ 54 $\pm$ 0.6\\ 55 $\pm$ 0.9 \\ 19 $\pm$ 0.0 \\ \tb{61} $\pm$ \tb{1.9}\end{tabular} & 
\begin{tabular}[c]{@{}c@{}}55 $\pm$ 0.5\\ 55 $\pm$ 0.4\\ 60 $\pm$ 1.1 \\ 30 $\pm$ 0.1 \\ \tb{79} $\pm$ \tb{4.3}\end{tabular} & 
\begin{tabular}[c]{@{}c@{}}81 $\pm$ 0.6\\ 81 $\pm$ 0.7\\ 81 $\pm$ 0.6 \\ 22 $\pm$ 0.0 \\ \tb{82} $\pm$ \tb{0.8} \end{tabular} & 
\begin{tabular}[c]{@{}c@{}}82 $\pm$ 0.3\\ 82 $\pm$ 0.3\\ 71 $\pm$ 1.3 \\ 32 $\pm$ 0.1 \\ \tb{88} $\pm$ \tb{1.5} \end{tabular} & 
\begin{tabular}[c]{@{}c@{}}81 $\pm$ 0.6\\ 81 $\pm$ 0.7\\ 68 $\pm$ 1.3 \\ 43 $\pm$ 0.0 \\ \tb{92} $\pm$ \tb{0.7} \end{tabular} & 
\begin{tabular}[c]{@{}l@{}}56 $\pm$ 0.3\\ 56 $\pm$ 0.3\\ 79 $\pm$ 2.0 \\ 48 $\pm$ 0.1 \\ \tb{94} $\pm$ \tb{1.1}\end{tabular}             \\ \hline
\end{tabular}
\end{adjustbox}
\end{table}

\vspace{-.3cm}
\section{Conclusions}
\vspace{-.2cm}

In this work we presented SAN, a new neural architecture that processes signals defined over simplicial complexes performing convolutional filtering over upper and lower neighborhoods using two independent masked self-attentional mechanisms. For an arbitrary $k$-simplex, the simplicial attention layer assigns different importances to different $(k\pm1)$-simplex' features that are within the lower and upper neighborhoods. The proposed architecture is also equipped with an harmonic filtering operation, which extracts relevant features from the harmonic component of the input data. In the experimental section, we have shown that in three different benchmarks, both inductive and transductive, SAN outperforms the current state-of-the-art architectures.

\bibliography{biblio.bib}

\end{document}